\documentclass[runningheads]{llncs}

\usepackage{url,subfigure,graphicx,amsmath,amssymb,listing,hyperref,doi}
\usepackage[font=small,labelfont=bf]{caption}
\usepackage[numbers]{natbib}
\usepackage{fancyvrb}

\DeclareMathOperator*{\Expect}{\mathbb{E}}

\author{James McDermott}

\institute{College of Engineering and Informatics,\\
  National University of Ireland, Galway.\\
  \url{james.mcdermott@nuigalway.ie}}

\title{When and Why Metaheuristics Researchers\\ Can Ignore ``No Free Lunch'' Theorems}
\titlerunning{When and Why Metaheuristics Researchers Can Ignore NFL}

\begin{document}

\maketitle

\footnotetext{This work was published in Springer Metaheuristics 2019 DOI 10.1007/s42257-019-00002-6: this journal is now discontinued.}

\begin{abstract}
  The No Free Lunch (NFL) theorem for search and optimisation states
  that averaged across all possible objective functions on a fixed
  search space, all search algorithms perform equally well. Several
  refined versions of the theorem find a similar outcome when
  averaging across smaller sets of functions. This paper argues that
  NFL results continue to be misunderstood by many researchers, and
  addresses this issue in several ways. Existing arguments against
  real-world implications of NFL results are collected and re-stated
  for accessibility, and new ones are added.  Specific
  misunderstandings extant in the literature are identified, with
  speculation as to how they may have arisen. This paper presents an
  argument against a common paraphrase of NFL findings -- that
  algorithms must be specialised to problem domains in order to do
  well -- after problematising the usually undefined term
  ``domain''. It provides novel concrete counter-examples illustrating
  cases where NFL theorems do not apply. In conclusion it offers a
  novel view of the real meaning of NFL, incorporating the anthropic
  principle and justifying the position that in many common situations
  researchers can ignore NFL.

\end{abstract}

\section{Introduction} \label{sec:introduction}



The ``No Free Lunch'' (NFL) theorems for search and optimisation are a
set of limiting results, stating that all black-box search and
optimisation algorithms have equal performance, averaged across all
possible objective functions on a fixed search space. 

NFL is famous: the 1995 technical report and the 1997 journal article
which introduced the original
results~\citep{wolpert1995no,wolpert-macready} have between them been
cited over 7000 times, according to Google
Scholar\footnote{\url{http://scholar.google.com/scholar?q='no+free+lunch'+wolpert+macready},
  20 February 2018.}, with 3000 of these since 2013.  They caused
considerable controversy when first published, and continue to divide
opinion. Some authors regard them as very important limiting results,
the equivalent of G\"odel's Incompleteness Theorem for search and
optimisation~\citep{ho-pepyne}. Others regard them as
trivial~\citep{haggstrom2007uniform, aaronson} or unimportant in
practice~\citep{hutter2010complete}. Still others,
including~\citeauthor{wolpert1995no} themselves, argue that the
practical importance of NFL is the implication that success in
metaheuristics requires matching the algorithm to the problem. Those
who argue that NFL is unimportant in practice do not dispute the
results {\em per se}. Many refinements have been added to the original
theorem, generally in the direction of ``sharpening'' it, i.e.~proving
new situations where NFL-like results hold. In this paper, we will use
the term ``the original NFL'' to refer to the original
theorem~\citep{wolpert1995no,wolpert-macready}, and ``NFL'' to refer
collectively to the original and refinements.


Proving NFL results requires some mathematics, but an understanding of
the theorems is not difficult. In fact, a simple paraphrase is
sufficient for discussion purposes, but it must be the {\em right}
paraphrase. Misinterpretations are common, both in the academic
literature and in technical discussion in blogs and web forums. Some
misinterpretattions will be examined in detail in later sections. Of
course, correct interpretations of NFL research are routinely given by
researchers who write specifically about the topic (as opposed to
those who mention NFL in the course of other work). However, the
literature tends not to directly take on and expose the source of
misunderstandings, nor provide concrete guidance. 


Therefore, this paper is aimed firstly at researchers who are left in
doubt by the existing NFL literature. It is intended to be a
``one-stop shop''. It argues broadly for the position that
metaheuristics researchers {\em can ignore NFL} in many common
situations. Rather than requiring researchers to state specific,
per-problem assumptions to escape NFL, or specialise their algorithms
to specific problem sets, it argues that existing algorithms may
already be specialised to an appropriate problem set, and provides a
single assumption which is well-justified and sufficient to escape NFL
for practical situations. Overall the paper is not focussed on new
research results, but rather a combination of an accessible review
with interpretation of results and consequences, new
results\footnote{A statement of practical consequences of more recent
  NFL variants; several corrections of NFL misunderstandings;
  problematising the term ``problem domain''; argument that existing
  generic algorithms are already specialised; fitness distance
  correlation and modularity as escapes from NFL; concrete NFL
  counter-examples in several domains; and introduction of the
  anthropic principle as a justification for the position that in many
  common situations researchers can ignore NFL.}, and a conclusion
which it is hoped is useful in practice.

Section~\ref{sec:review} summarises the NFL literature, concluding
that the strongest versions of NFL do not have stronger practical
implications than more well-known older versions.
Section~\ref{sec:misund-theor} describes many common NFL
misunderstandings, and in particular argues that existing, generic
algorithms are {\em already specialised} to an appropriate subset of
problems, potentially escaping NFL. The next two sections provide
specific assumptions which allow researchers to ``escape'' NFL in
practice (Section~\ref{sec:structure}) and concrete counter-examples
(Section~\ref{sec:examples}). But Section~\ref{sec:discussion} goes
further to explain why, based on the anthropic principle, researchers
can in many situations ignore NFL without making specific, per-problem
assumptions. Section~\ref{sec:conclusions} notes that common ``rules
of thumb'' about metaheuristic performance may remain true even though
NFL does not support them, and summarises by stating when and why
metaheuristics researchers can ignore NFL.

\section{Review} \label{sec:review}

In many fields of research, a certain type of limiting result is
sometimes given the nickname ``no free lunch''. There are two strands
of NFL research of interest here: NFL for search and
optimisation~\citep{wolpert1995no, wolpert-macready}, the main topic
of this paper, and NFL for supervised machine
learning~\citep{wolpert1996lack}, which will be briefly discussed in
Section~\ref{sec:other-nfl} and Section~\ref{sec:conclusions}. NFL
results in other fields such as reinforcement learning, physics or
biology are not treated in this paper.

In this section we review the original NFL for search and optimisation
and several refinements, mostly in chronological order, discuss the
biggest contributions of NFL research, and then summarise.  We remark
that several good reviews of the NFL literature have been presented
elsewhere, including those by~\citet{corne2003no,
  whitley2005complexity, joyce2018review}.

\subsection{NFL for search and optimisation}

The (original) NFL theorem for search and
optimisation~\citep{wolpert1995no, wolpert-macready} states that the
performance of any two deterministic non-repeating black-box search
algorithms is equal, regardless of the performance measure, when
averaged across all possible objective functions on a given search
space. That is:
$$\sum_{f\in F} P(d^y_m | f, m, a_1) = \sum_{f\in F} P(d^y_m | f, m, a_2)$$
\noindent where $a_1, a_2$ are search algorithms, $F$ is the set of
all possible functions on a given search space, $P$ is a performance
measure, and $d_m^y$ is a {\em trace}, i.e.~the history of objective
function values of individuals visited by the algorithm in its first
$m$ steps. A performance measure is a single number reflecting how
well an algorithm has done, given a set of objective function values:
for example, ``best ever objective value'' is a performance measure,
and ``mean objective value of the trace'' is another. Finally, a
search algorithm is a procedure which chooses which point in a search
space to visit next, given a trace. ``Black-box'' means that the
algorithm is given {\em only} a trace. It cannot ``see inside'' the
objective function, e.g.~by using its gradient. Such algorithms
include genetic algorithms, simulated annealing, particle swarm
optimisation, and many other metaheuristics.

The original NFL results are stated for deterministic algorithms, but
the stochastic case reduces immediately to the deterministic case and
the difference is never
crucial~\cite{wolpert-macready,joyce2018review}. NFL results are often
stated for non-repeating algorithms (a repeating algorithm can be
systematically worse than random search because it wastes
time~\cite{poli2009there}), but again this point is not important in
discussion of NFL, since any repeating algorithm can be made
non-repeating with memoisation~\cite{droste2002optimization}.

An immediate corollary of NFL is that the performance of every
algorithm is equal to that of random search. Better than random
performance on some functions is balanced by worse than random on
others. As \citet{koppen2001remarks} write, ``all attempts to claim
universal applicability of any algorithm must be
fruitless''. \citet{oltean2004searching} argued that ``these
breakthrough theories should have produced dramatic changes'' in the
field. However, \citet{wolpert-macready, radcliffe1995fundamental,
  wolpert2012no} and others discouraged ``nihilistic''
interpretations, instead arguing that the true meaning of NFL is that
an algorithm should be tailored to the problem in order to achieve
better than random performance.

Other researchers, while accepting NFL as stated, argued that NFL can
be dismissed because the scenario presented -- wishing to optimise all
possible functions -- is not realistic~\citep{droste1998perhaps}.
\citet{hutter2010complete} goes so far as to call NFL a ``myth'' with
little practical implication, explaining that ``we usually do not care
about the maximum of white noise functions, but functions that appear
in real-world problems''. \citeauthor{hutter2010complete} is making a
distinction between problems we care about and ones we do not. In the
context of machine learning, \citet{bengio2007scaling} similarly
define the ``AI-set'' as ``the set of all the tasks that an
intelligent agent should be able to learn''. In the same way, we will
use the term ``problems of interest'' to refer to optimisation
problems which we would like to be able to solve using optimisation
algorithms. This definition, like those of
\citeauthor{hutter2010complete} and \citeauthor{bengio2007scaling}, is
informal but sufficient for discussion. A first question for any NFL
investigation is then whether the set of problems of interest on a
given search space is equal to the set of all possible functions on
that space, since if not, the original NFL will not apply. A
fundamental objection to NFL is that the set of all possible functions
is enormous, and contains many ``unnatural'' functions which would
never arise in real
problems~(e.g.~\citet{droste1998perhaps}). However, we next consider a
variant of NFL which applies to a much smaller set.



\subsection{Sharpened NFL} \label{sec:sharpened-nfl}

The ``sharpened'' NFL (SNFL) theorem~\citep{schumacher2001no} can be
seen as a response to~\citet{droste1998perhaps}, because it states
that equal performance of algorithms holds over a much smaller set
than ``all possible objective functions'',
which~\citet{droste1998perhaps} had argued was unrealistic. The
statement of SNFL is the same as that of NFL, except that now
performance is averaged across any set of objective functions which is
{\em closed under permutation} (CUP). A CUP set is a set of objective
functions which differ from each other only by permuting the objective
function values -- i.e.~under any function in the set of functions,
the same objective values are allocated, but to different points of
the search space. Thus, given a search space, a CUP set is
characterised by a multi-set of objective values. Such a set is called
CUP because permutation does not result in a new function outside the
set.  For example, if $f$ assigns a unique value to each point $x\in
X$, then there are $|X|!$ functions in the CUP set of $f$. For another
example, consider the Onemax problem on bitstrings of length $n$:
$f(x) = \sum_i^n x_i$, and $f$ is to be maximised. There is just one
point $x' = (0, 0, \ldots, 0)$ which has $f(x')=0$ and is the worst
point in the space. Any function $f'$ in the CUP set of $f$ must also
award an objective value $f'(x'') = 0$ to one and only one point $x''$
in the space (not necessarily the same one). If there are multiple
points with an objective value $f' = 0$, then $f'$ is not in the CUP
set.

The set of all possible functions can be partitioned into CUP sets --
the CUP sets are disjoint and collectively exhaustive. SNFL holds over
each CUP set, so NFL can be derived as an immediate corollary of SNFL.


Although SNFL is ``sharper'' than NFL, in practice this sharpness also
gives researchers an easier way to ``escape'' NFL. Because SNFL is
``if and only if'' (the set of problems is CUP if and only if average
performance for all algorithms is equal), a proof that the set of
problems of interest is not CUP is sufficient to escape SNFL and thus
also escape the original NFL.

With this in mind, \citet{igel2004no} asked: ``are the preconditions
of the NFL-theorems ever fulfilled in practice? How likely is it that
a randomly chosen subset is [CUP]?'' The implicit assumption here is
that our set of problems of interest will be drawn uniformly from all
subsets, and according to the authors, the number of CUP subsets ``can
be neglected compared to the total number of possible subsets'', and
thus SNFL is unlikely to apply to our set of problems of
interest. However, as~\citet{whitley2005complexity} state, ``the {\em
  a priori} probability of any subset of problems is vanishingly small
-- including any set of applications we might wish to consider''. It
thus falls into much the same trap as NFL itself -- there are many
possible subsets, but most subsets are rather ``unnatural''. There is
no reason to think that the problems of interest are drawn uniformly
in this way. \citet{rowe2009reinterpreting} argue that using the
language of probability in this way is indeed misleading.

However, other researchers did indeed take advantage of the ``if and
only if'' nature of SNFL.  \citet{igel2001classes,
  koehler2007conditions} and others demonstrated that common
problem types, such as TSP, are not CUP, and hence on these neither
NFL nor SNFL constrains algorithm performance.

\citet{wegener2004} also stated that the SNFL scenario is not
realistic: ``We never optimize a function without: a polynomial-time
evaluation algorithm $(a,f) \rightarrow f(a)$; a short description;
structure on the search space''. Whether the evaluation algorithm is
polynomial-time is irrelevant in contexts where the search space is of
fixed size, but the broader point stands: CUP sets may include many
functions which require more than polynomial time to evaluate, or a
long description (e.g., no shorter than the table of objective
function values), which is equivalent to there being no structure on
the search space. The conclusion is that the set of functions of
realistic interest will never form a CUP set, and hence algorithms are
free to out-perform random search on them. Although some aspects of
this conclusion were later questioned (see
Section~\ref{sec:compressibility}), as a whole it probably stands as
the settled position of many researchers. \citeauthor{wegener2004}
considered this to be the last word: ``The NFL theorem is fundamental
and everything has been said on it [\ldots] It is time to stop the
discussion''. However, more was to come: firstly in the form of NFL
refinements, and secondly in misunderstandings of NFL results which
prevent the discussion from being closed.

\subsection{Non-uniform NFL} \label{sec:non-uniform-nfl}


\citet{igel2004no} argued that a simple {\em averaging} of performance
across all functions on the space (as envisaged in NFL), or the CUP
set (as in SNFL) is not relevant, since in practice problems may be
encountered according to a non-uniform probability distribution.

Previous to this, the idea that we ``care'' only about some of the
possible problems on a search space, and regard others as unimportant,
was treated as an all-or-nothing proposition: for each problem, we
either care about it or we do not. It was also a route to avoiding
NFL, since when only a subset of functions are of interest, an
algorithm is free to out-perform random search. Using a non-uniform
distribution generalises this idea, so that we may say we care about
different problems to different degrees, weighting them according to a
distribution. The non-uniform NFL theorem (NUNFL) theorem proved
independently by
\citet{streeter2003two,igel2004no,english2004structure} and (according
to~\citet{english2004no})~\citet{neil-woodward} clarifies the
implications. According to NUNFL, all algorithms perform the same when
taking their weighted mean performance over any set of functions $F$,
where weighting is according to some probability distribution, if and
only if the distribution is {\em constant} on any CUP set within $F$,
also known as a ``block-uniform'' probability distributions on
problems.

This result does not greatly change the overall NFL picture. Just as
before, if we can show or assume that in a given CUP set, some
functions are of interest and some are not, then the probability
distribution is not constant on that set (not block-uniform), and so
no NFL, SNFL or NUNFL result holds. SNFL can be seen as a corollary of
NUNFL.

As discussed above, \citet{igel2004no} argued that the probability of
a set (e.g., the set of problems of interest) being CUP is vanishingly
small, but went on to discard this argument. However, they use a very
similar argument in the context of NUNFL: ``The probability that a
randomly chosen distribution over the set of objective functions
fulfills the preconditions of [NUNFL] has measure zero. This means
that in this general and realistic scenario the probability that the
conditions for a NFL-result hold vanishes.'' In fact, this scenario is
still not realistic, for much the same reasons: our distribution over
problems is not chosen uniformly from all possible distributions.

\subsection{Focussed NFL and restricted metric NFL} \label{sec:focussed-nfl}

\citet{whitley2008focused} further refined NFL to produce ``focussed''
NFL (FNFL). Where SNFL states that {\em all} algorithms perform
equally on a CUP set, FNFL states that for any given subset {\em of
  algorithms} there is a closure set of functions (possibly much
smaller than the CUP set) over which the algorithms perform
equally. This new closure set, called the ``focus set'', is derived
from the orbits (components) of permutations representing the
behaviour of the given algorithms on any function. The most extreme
example discussed by the authors concerns just a pair of algorithms
$(A_1, A_2)$ and a single function $f_1$. Running $A_1$ on $f_1$ gives
a trace $T$. Using permutations representing $f_1$ and $T$, we can
construct a function $f_2$ such that running $A_2$ on $f_2$ will give
the same trace $T$, so the focus set is $\{f_1, f_2\}$. With identical
traces, any performance measure will be identical. In the example,
both $f_1$ and $f_2$ are ``toy problems'', not of practical interest.

\citet{joyce2018review} went on to produce a yet sharper result, the
restricted metric NFL (RMNFL). Again we start with a restricted set of
algorithms and of functions, but now also a restricted set of
performance metrics. Given a set of functions $\beta$, RMNFL tells us
that there exists a ``restricted set'' on which all of the considered
algorithms have equivalent performance according to the considered
performance metrics, and this restricted set is a subset (and may be a
proper subset) of the focus set. Again the example considers ``toy
problems''.

\begin{figure}
  \centering
  \includegraphics[scale=0.65]{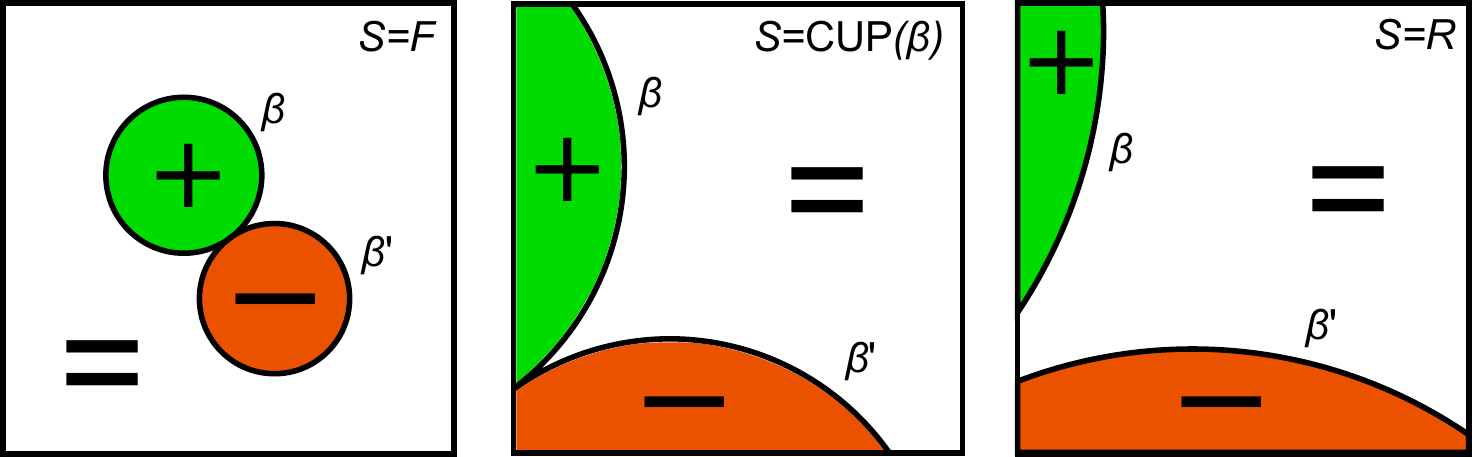}
  \caption{Schematic representing performance of an algorithm $A$
    relative to random search. Left: by the original NFL, if
    performance of $A$ over set $\beta$ of objective functions on a
    fixed search space is better than random ({\tt +}), then it is
    worse than random on the remainder, which may be composed of a
    fraction $\beta'$ where it is truly worse than random ({\tt -})
    and a fraction $F-\beta-\beta'$ where it is equivalent to random
    ({\tt =}). Centre: choosing a new $\beta$, by SNFL, the same is
    true over a CUP set. Right: choosing a new $\beta$, by RMNFL, the
    same is true over the ``restricted'' set. The existence of
    problems on which RS out-performs $A$ is already guaranteed by
    SNFL. We can think of RMNFL as zooming-in on SNFL, and SNFL as
    zooming-in on the original NFL. Although RMNFL is stronger than
    SNFL, the practical implications seem the
    same.\label{fig:nfl-snfl-rmnfl}}
\end{figure}

One useful strategy for understanding the practical implications of
NFL results is to consider the performance of our favourite algorithm
$A$ in comparison with that of random search (RS). The original NFL
theorem and several refinements SNFL, FNFL and RMNFL then all lead to
similar-sounding remarks: for a fixed search space, if $A$
out-performs RS averaged over a set of problems $\beta$ then there
exists a set of problems $S$ such that $\beta \subset S$ and RS
out-performs $A$ on the remainder $S-\beta$. It's important to realize
that ``RS out-performs $A$ on the remainder'' does not necessarily
mean that RS out-performs $A$ on every single problem in $S-\beta$,
but rather that it out-performs $A$ averaged over $S-\beta$. It may be
that RS out-performs $A$ on another subset $\beta'$, and they have
equivalent performance on $S-\beta-\beta'$, as illustrated in
Fig.~\ref{fig:nfl-snfl-rmnfl} (left).

The difference between NFL and the various refinements is the identity
of $S$, as shown in Fig.~\ref{fig:nfl-snfl-rmnfl}. In NFL, $S=F$, the
set of all possible problems on the search space. In SNFL, $S =
\mathrm{CUP}(\beta)$, the CUP set. In FNFL, $S=\Phi$, where $\Phi$ is
the ``focus set'' constructed with reference to our set of algorithms
($A$ and RS), and $\Phi \subseteq \mathrm{CUP}(\beta)$. In RMNFL, $S =
R$, where $R$ is the ``restricted set'' constructed with reference
both to the set of algorithms and the choice of performance metrics,
and now $R \subseteq \Phi$ (and it may be that $R=\Phi$). Overall, then:

$$\beta \subset R \subseteq \Phi \subseteq \mathrm{CUP}(\beta) \subset F$$

Informally, it is useful to think of three types of objective function
on any search space. ``Nice'' functions give ``the right
hints''~\citep{droste2002optimization} to the algorithm. There are
also deceptive ones, which have similar types of structure but give
misleading hints. Finally there are the random functions, which have
effectively no structure and are by far the most numerous. Thus, in
Fig.~\ref{fig:nfl-snfl-rmnfl} (left and centre), the sets marked {\tt
  +} and {\tt -} are very small relative to that marked {\tt =}.

The sets $R$ and $\Phi$ can be much smaller than $\mathrm{CUP}(\beta)$
which is itself usually far smaller than $F$, and so an intuitive
argument researchers may use to ``escape'' NFL and SNFL (``the set of
problems considered by NFL or SNFL is so huge that it likely contains
some pathological problems which are unimportant in practice'') cannot
apply to escape FNFL or RMNFL. However, the practical implications are
the same: FNFL and RMNFL guarantee the existence of problems where RS
out-performs $A$, and we already knew this from SNFL. These problems
can only be a subset of the ones already identified by SNFL. FNFL and
RMNFL don't show that these problems (where RS out-performs $A$) are
of interest. They may provide a {\em mechanism} by which researchers
could in principle show this, but this has not been done for any
example yet, to our knowledge. Thus, researchers can escape FNFL and
RMNFL by the same assumption as they escape NFL and
SNFL~\cite{hutter2010complete}: there are problems where $A$ performs
worse than RS, but it is assumed that they are not of interest. The
practical implications of FNFL and RMNFL are no stronger than those of
SNFL.

Moreover, even where a focus set or restricted set exists as a proper
subset of the set of problems of interest to a researcher, if that set
of problems of interest is not CUP, then SNFL still applies in the
``and only if'' direction: algorithm $A$ can out-perform RS on it.






\subsection{NFL in the continuum} \label{sec:continuum}

Another relatively recent development is the application of NFL theory
in continuous search spaces, in contrast to the discrete spaces
considered in most NFL literature. \citet{auger2010continuous} claim
that NFL does not hold in continuous spaces,
but~\citet{rowe2009reinterpreting} argue that this result occurs only
because of an incorrect framing of the problem in probabilistic
language which generalises with difficulty to the
continuum. \citeauthor{rowe2009reinterpreting} show that an NFL-like
result does indeed hold. \citet{alabert2015no} agree with and build
on~\citeauthor{auger2010continuous}, without
citing~\citeauthor{rowe2009reinterpreting}. We will not enter into
this debate, partly because in practice metaheuristic algorithms do
run in an effectively discrete setting~\cite{joyce2018review}.

\subsection{Reinterpreting and re-proving NFL} \label{sec:reinterpreting}

Several authors have given re-interpretations of NFL which are perhaps
more intuitive than the original.

\citet{culberson1998futility} remarks that choosing an objective
function $f$ uniformly is equivalent to choosing each objective value
$f(x)$ uniformly from the objective set (e.g.~a subset of
$\mathbb{R}$) at the point in time when the algorithm first chooses to
visit $x$. With this ``adversarial'' view, it perhaps becomes easier
to intuit just how hopeless black-box search is over ``all possible
fitness functions''. \citet{haggstrom2007uniform} also uses this type
of reasoning to give a good phrasing of NFL. \citet{serafino2013no}
states a ``no induction form'' of NFL: with no prior knowledge on the
objective function, each objective function value provides no
information on any other point in the search space.
\citet{woodward2003no} point out that by NFL-like reasoning, every
algorithm will visit the optimum {\em last} on some
function. \citeauthor{culberson1998futility} makes his point in
relation to the original formulation of NFL, but the same reasoning
can be applied to the SNFL formulation: choosing a objective function
$f$ uniformly from the CUP set is equivalent to choosing each
objective value $f(x)$ uniformly {\em from those so far unused in the
  multiset}. \citet{joyce2018review} improve on this reasoning to give
a new approach to proving NFL results. It uses the {\em trace tree}
representation of search algorithms, introduced
by~\citet{english2004no}. In the trace tree, each node is labelled
with the ``trace so far'' (traversing from the root), and it branches
each time the algorithm observes an objective value and makes a
decision about which point to visit next. A simple counting argument
given by~\citet{joyce2018review} shows that on a fixed search space,
all algorithms produce the same set of traces.

\citet{rowe2009reinterpreting} ``reinterpret'' NFL focussing on
symmetry results rather than applications, and removing the language
of probability which they argue is inappropriate. As they point out,
it is part of a general trend in NFL thinking, also present in works
by~\citet{english2004no, duenez-guzman_vose, joyce2018review} and
others, towards abstraction, duality between algorithms and functions,
and use of permutations and eventually group actions to represent
behaviour.

\subsection{Special situations}

There are several special situations where NFL results do not apply.

In {\bf multi-objective optimisation} (MOO), the objective function is
vector-valued, e.g.~both maximising the strength and minimising the
weight of an engineering design. NFL applies to such
scenarios. However,~\citet{corne2003no} remark that in MOO it is
common to use {\em comparative} performance measures, i.e.~measures
where we never consider the performance of an algorithm, but only the
performance of one relative to another. This prevents an NFL result
because although (as reasoned in SNFL) every possible trace will occur
as the trace of every algorithm on some problem, a given {\em pair} of
traces generated by two algorithms may not occur by any other
pair of algorithms for any problems.

In {\bf coevolution}, the objective function of an individual in the
population is defined through a contest between it and others. If a
search algorithm allows the ``champion'' individual to contest only
against very weak antagonists then it is wasting time, in a way
somewhat analogous to a repeating algorithm, and so can be
systematically worse than random
search~\citep{wolpert2005coevolutionary}.

In {\bf hyper-heuristics}, the idea is to use a metaheuristic to
search for a heuristic suitable to a problem or set of problems. In
typical hyper-heuristic scenarios, the set of problems on which the
heuristic is evaluated must be small, in order for search to be
practical, and in particular small relative to the number of possible
objective function values in a typical scenario. \citet{poli2009there}
argue that therefore the higher-level problem (from the point of view
of the hyper-heuristic) cannot be CUP, so SNFL does not apply.

These cases will not be considered further here.



\subsection{The contributions of NFL} \label{sec:contribution}

Although we will argue that researchers are in most situations
justified in ignoring NFL, it has made important contributions to the
field of metaheuristics.

\begin{figure}
  \centering
  \includegraphics[width=0.6\linewidth]{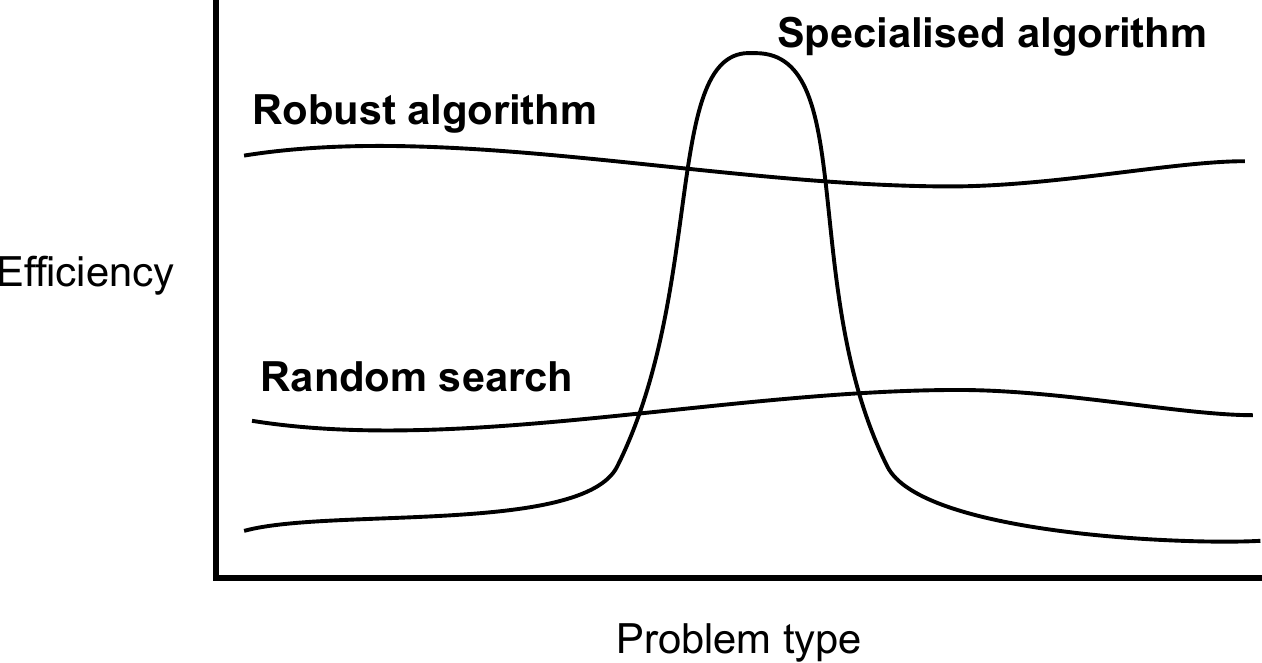}
  \caption{An incorrect intuition on algorithm performance (adapted
    from~\citet{goldberg})\label{fig:goldberg}}
\end{figure}

The first contribution is that NFL corrects a vague intuition which
was perhaps common prior to NFL, illustrated in
Fig.~\ref{fig:goldberg}. It purports to show that while a specialised
algorithm can be the best on the problems it is specialised to, a
``robust'' algorithm such as a genetic algorithm can hope to be better
than random search on all problems. If this ``all'' is taken
literally, then this is incorrect and nowadays researchers making such
claims would be marked as cranks. NFL is thus the equivalent of the
proof in physics of no perpetual motion machines.


A related benefit of NFL is that it reminds us to be careful, when
making statements of the form ``algorithm $A$ out-performs algorithm
$B$'', to specify what set of problems we are considering. For ``all
possible problems'' or ``all problems in a CUP set'', or a focus set
or restricted set, the statement is false; for ``all problems of
interest'' the statement is an almost incredibly strong claim, but not
disallowed by NFL; and so most such statements should be for sets
identified in other ways. Whenever an algorithm $A$ out-performs
random search on a set of problems, it is because $A$ is specialised
to the set of problems, and NFL encourages us to identify this
specialisation, as discussed in Section~\ref{sec:structure}.

Finally, NFL results have stimulated a greatly improved understanding
of search algorithms. This includes symmetry results and a view of
algorithms and problems as permutations and eventually as group
actions~\citep{rowe2009reinterpreting}; a view of algorithm behaviour
through the lens of decision trees~\citep{english2004no,
  joyce2018review}; and an important connection between NFL and
Bayesian optimisation. After observing a (partial) trace of objective
function values, an optimisation algorithm is in a position to update
its prior on the location of the optimum. But in order to do so it
requires also a prior on the distribution of objective
functions~\citep{serafino2013no}.

\subsection{Summary}
\label{sec:how-can-performance}




The original NFL and its variants seem never to have been applied to
demonstrate a limit on performance in practice. It is more common to
respond to NFL by exhibiting functions (typically, pathological ones
on which our algorithm will do badly) and claim that we do not care
about performance on them, and thus performance on problems we {\em
  do} care about is free to be better than random. One example
is~\cite{hutter2010complete}, referring to ``white noise
functions''. With SNFL, discussion often hinges on taking a set of
problems on a given search space, and asking whether it is CUP or
not. This means asking whether any permutation of objective values of
a function in that set will result in a new function in the same
set. Importantly, SNFL is an ``if and only if'' result. It is common
to show that some set of problems of interest is not CUP, so at least
according to SNFL algorithms are free to perform differently: examples
include~\cite{igel2001classes, koehler2007conditions, poli2009free,
  poli2009there}, as discussed in detail in
Section~\ref{sec:structure}. The secondary literature has not yet
attempted to make sense of the practical implications of FNFL and
RNFL.



\section{Misunderstanding NFL} \label{sec:misund-theor}

\citet{woodward2003no, sewell2012forecasting} both state that NFL is
often misunderstood; \citet{wolpert2012no} writes that much research
has arguably ``missed the most important implications of the
theorems''. In this section, several actual and potential
misinterpretations will be presented. We can begin by noting that the
colloquial sense of the phrase ``no free lunch'' should not be
confused with the technical sense, as done
e.g.~by~\citet{lipson-nutonian}.





\subsection{Going beyond a single search space or problem size}
\label{sec:going-beyond-single}
\label{sec:scalable-problems}

If we observe good performance of a given algorithm on several
problems from a given problem domain (e.g.~TSP), then a natural
response, given the overall thrust of NFL, is to expect bad
performance on another domain (e.g.~symbolic regression). However,
since the search space for symbolic regression is different from that
for TSP, NFL gives no indication of performance on it.

Similarly, it may be natural to think that, given an observation of an
algorithm performing well on some problem sizes, it will necessarily
perform badly on other problem sizes. For example,
\citet{watson1999algorithm} find that better-than-random performance
on several synthetic problems fails to transfer to real-world
problems, and introduce NFL to discuss the reasons why. This is
inappropriate since their synthetic and real-world problems have
different sizes, hence different search spaces. In practice there may
often be reasons to believe that good performance on a problem of a
given size tends to indicate {\em good} performance on the same
problem at a different size. These reasons are independent of NFL.

It might be countered that we can define a single large search space
to include embeddings of multiple search spaces, with different sizes
of each, and that NFL would apply to this space. This is correct:
however, our algorithm does not run on this space. This argument
constrains the performance only of algorithms designed to run on that
space.

These misunderstandings may arise from a bad paraphrase of the
original NFL, along the lines of ``you can't win on all
problems''. It's important to include ``\ldots on a fixed search
space''.

\subsection{Superfluous references} \label{sec:superfluous-use}

As \citet{koppen2001remarks} remark, as a reaction to the original
NFL, many references ``were put into the introductory parts of
conference papers''. Perhaps many researchers take the view that
claims about overall algorithm performance could be criticised by a
reviewer citing NFL, so it is safer to put in a pre-emptive, defensive
NFL reference.  Such statements are not incorrect, but are often
superfluous. An example is chosen arbitrarily from a highly-cited
paper in a top journal: ``Simulation results showed that CRO is very
competitive with the few existing successful metaheuristics, having
outperformed them in some cases, and CRO achieved the best performance
in the real-world problem. Moreover, with the No-Free-Lunch theorem,
CRO must have equal performance as the others on average, but it can
outperform all other metaheuristics when matched to the right problem
type''~\citep{lam2010chemical}. The behaviour is damaging when the
algorithm in question is of the type criticised
by~\citet{sorensen2015metaheuristics} and \citet{weyland2015critical} --
algorithms of dubious novelty disguised by far-fetched nature-inspired
metaphors. Sometimes the authors of such algorithms cite NFL
superfluously, perhaps as a show of respectability,
e.g.~\citep{al2012harmony, ouaarab2014discrete,
  chawda2017investigating}, and \citep[][p.~19]{beheshti2014capso}. It
is not only authors who are at fault here: they may be correct in
guessing that reviewers will make spurious references to NFL, so the
behaviour is incentivised.

\subsection{False intuition concerning the size of NFL-relevant sets}
\label{misunderstanding-tsp-cup}
\label{sec:valid-from-small}

By the original NFL all algorithms have equal performance over the set
$F$ of all possible functions $f$ on a fixed search space $X$. It is
easy to underestimate just how large $F$ is. The same applies to ``all
functions in a CUP set'' in SNFL.

For example, if we are working on genetic programming (GP) symbolic
regression~\citep{koza}, the phrase ``all possible functions'' will
naturally bring to mind ``all possible regression problems'', perhaps
represented by a phrase like ``all possible training sets''. But this
set is not the same as ``all possible problems on the regression
search space'', which includes many problems which are not regression
problems for any dataset. NFL does not apply to the far smaller and
better-behaved set of all symbolic regression problems (see
Section~\ref{sec:symb-regr-not}).

Similarly, when discussing TSP problems, which are defined on a search
space itself consisting of permutations, it is easy to confuse ``all
TSPs'' with ``all problems on the permutation search space'' or ``all
problems in the CUP set of a TSP''. These are not the same.
\citet{woodward2003no} write: ``From a given scenario [i.e.~a TSP
  instance] other problems can be generated simply by re-labelling the
cities. If there are $n$ cities, there are $n!$ ways of re-labelling
these cities (i.e.~all permutations)''. They conclude that NFL holds
for TSP. This confuses permutations of cities with permutations of
objective values. 

Several authors have found results where different algorithms win on
different problems, and claimed this as validation of or evidence for
NFL. For example, \citet{ciuffo2014no} write that ``[t]he performance
of the different algorithms over the 42 experiments considerably
differ. This proves the validity of NFL theorems in our field''
(traffic simulation). \citet{Vrugt708} claim that their results for
several algorithms ``provide numerical evidence [of NFL], showing that
it is impossible to develop a single search algorithm that will always
be superior to any other algorithm''. These statements are
inappropriate as the problem sets considered are far too small to be
bound by NFL or SNFL and there is no claim concerning the more refined
variants FNFL and RNFL. \citet{oltean2004searching} uses evolutionary
search to find problems on which one given algorithm is out-performed
by another. This helps to illustrate NFL, but is not evidence of NFL's
truth or otherwise.

For another example, \citet{aaronson} paraphrases NFL as ``you can't
outperform brute-force search on random instances of an optimization
problem''. The phrasing is not correct: for any optimisation problem,
the set of instances of that problem is far smaller than the set of
objective functions on that problem's search space.

Finally, \citet{wolpert2012no} writes: ``the years of research into
the traveling salesman problem (TSP) have (presumably) resulted in
algorithms aligned with the implicit $P(f)$ describing traveling
salesman problems of interest to TSP researchers.'' Here, $P$ is a
probability distribution over TSP problems, so this statement is in
the framework of NUNFL. But since the set of all TSP problems is far
smaller than the set of all possible problems on the same space, and
is not CUP, it is entirely possible (at least, according to NFL) for
TSP researchers' algorithms to be uniformly excellent across all TSP
problems, whether in a uniform or any other distribution.

Of course, in FNFL and RMNFL the problem sets on which no algorithm
out-performs RS may be far smaller than the set of all possible
functions or a CUP set. But the constructions provided
by~\citet{whitley2008focused} (for FNFL) and~\citet{joyce2018review}
(for RMNFL) do not lead to an easy intuition on what problems in the
focus set or restricted set are like. Any intuition that because they
are smaller than the CUP set, they tend not to contain ``ill-behaved''
functions is not supported by the evidence so far.

\subsection{The undefined terms ``problem domain'' and ``problem-specific''}
\label{sec:undef-terms-probl}

As already stated, a common interpretation of NFL is that
``problem-specific knowledge'', ``domain knowledge'' or ``information
on the problem class'' is required to be embodied in the algorithm in
order to out-perform random search. For example,
\citet{wolpert-macready} write that NFL results ``indicate the
importance of incorporating problem-specific knowledge into the
behavior of the algorithm''. These statements can be easily
misinterpreted, because terms such as ``problem domain'' and
``problem-specific'' are ambiguous.

The mathematics of NFL does not deal with application domains, only
problem subsets. For example, \citet{radcliffe1995fundamental}
introduce terminology with the following phrasing: ``Let $D$ be a
(proper) subset of $R^S$ (a problem ``domain'').'' Here $R^S$ is the
set of all possible objective functions on the search space $R$. The
scare quotes around {\em domain} hint at the problem. When we read the
word ``domain'', we do not think of an arbitrary subset of the set of
all possible objective functions: we think of an {\em application
  domain}, i.e.~a set of problems in a particular application area,
such as vehicle routing problems, or bioinformatics. It is better to
avoid the term ``domain knowledge'' and think instead of a term like
``problem subset knowledge'', to avoid smuggling in false
connotations. Thus we may say that to out-perform random search on a
subset of all possible problems, we require knowledge of that subset.

``Problem subset knowledge'' must mean something like ``knowledge that
a given property holds for problems in the subset, but not for
problems outside it''. In particular, ``problem-specific knowledge''
might be taken to mean that the property in question is true {\em
  only} for that problem and no others. For example, \citet{ho-pepyne}
write that: ``the only way one strategy can outperform another is if
it is specialized to the {\em specific} problem under consideration''
(emphasis added). But a hill-climber out-performs random search on the
Onemax problem, and no-one would say that hill-climbing is specialized
{\em only} to Onemax, or that it takes advantage of some property of
Onemax which is not true of any other problem. More broadly, it is
possible to out-perform random search on a problem subset {\em on
  average} by taking advantage of a property which is {\em common but
  not universal} in that subset. Of course,
as~\citet{wolpert-macready} write, it is not enough for the creator of
an algorithm to be aware of any such property -- it must be embodied
in the algorithm. Exactly what types of knowledge might be used, and
how, are discussed in Section~\ref{sec:tsp-not-cup}.

\subsection{Tailoring the algorithm to the problem}
\label{sec:algorithm-must-be}

``Hammers contain information about the distribution of nail-driving
problems'' -- \citet{english1996evaluation}.

A common interpretation of NFL is given
by~\citet{whitley2005complexity}: algorithms must be tailored to
problems, and thus ``the business of developing search algorithms is
one of building special-purpose methods to solve application-specific
problems.'' This suggests that when we encounter new problems, we
should seek to understand their properties and design specialised
algorithms for them. This is also the position
of~\citet{wolpert-macready}
and~\citet{wolpert2012no}. \citet{smith2009cross} argues (based on
\citet{rice1976algorithm} and on NFL) that we should ``move away from
a `black-box' approach to algorithm selection, and [\ldots] develop
greater understanding of the characteristics of the problem in order
to select the most appropriate algorithm'', recommending landscape
analysis and other methods to develop such understanding. This
position represents common and good practice, independent of NFL.

\begin{figure}[h!]
  \centering
  \includegraphics[scale=0.5]{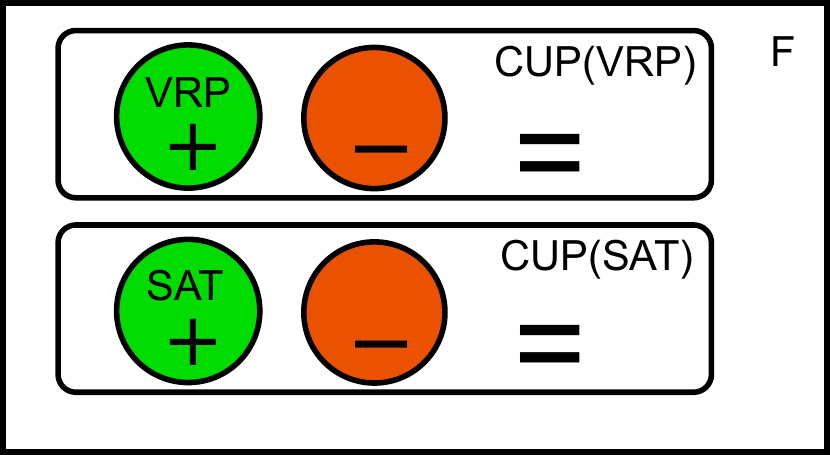}
  \caption{One possible scenario for performance on a set $F$ of
    problems of fixed size, not disallowed by any NFL result. Here
    CUP(VRP) indicates the CUP set of a particular VRP instance. If
    for any application domain its instances always occur as a proper
    subset of a CUP set, then SNFL allows a single algorithm to
    out-perform random search ({\tt +}) on {\em all} application
    domains, balanced by worse-than-random ({\tt -}) and
    equal-to-random ({\tt =}) performance within each CUP set on
    problems which are not of
    interest.\label{fig:one-algo-to-rule-them-all}}
\end{figure}

As already argued, the set of all problems on a fixed search space
does not necessarily correspond to an application domain. Neither need
each application domain correspond to a CUP set (SNFL), a focus set
(FNFL), or a restricted set (RMNFL). For example, if we observe an
algorithm having better-than-random performance on a set of VRP
instances of fixed size (hence it is not a CUP set), then there is
nothing in SNFL to prevent better-than-random performance on another
set of problems on the same space, such as instances of Satisfiability
(SAT). The scenario is illustrated with respect to SNFL in
Fig.~\ref{fig:one-algo-to-rule-them-all}.

Crucially, no assumption is required here that is not already made by
those who advocate incorporating problem-specific knowledge. But the
conclusion is very different: researchers can feel free to try out
existing algorithms on new problems, and even to search for new
``super algorithms'' which are better than existing algorithms
averaged across all problems of interest (nevertheless, such claims
would require precise formulation and extraordinary evidence).

In fact, although it is unlikely, this scenario could already be the
case with no contradiction to NFL results, for some algorithm such as
a genetic algorithm, or stochastic hill-climbing with restarts. If so,
then by NFL, these generic algorithms must be {\em already
  specialised} to some problem subset. This sounds like a
contradiction in terms. Our crucial claim is that generic algorithms
are specialised to a particular set of problems which we might
characterise as problems of interest. An algorithm can achieve better
than random performance averaged across a subset of problems (not
necessarily on every single one) even if no-one has given a formal
definition for that subset, and we do not intend to attempt a
definition of the set of problems of interest here, though we discuss
one way to think about it in
Section~\ref{sec:real-meaning-nfl}. (\citet{schumacher2001no} state
that ``all algorithms are equally specialized'', just to different
sets of functions, but this relies on also viewing random search as
specialised, by considering it as a deterministic algorithm with the
random seed as a parameter.)


The specialisation to this large, ill-defined subset has happened
through intuition, trial and error, theoretical understanding of the
properties of the problems of interest, and gradual matching of their
properties in algorithms. Algorithms are {\em designed} according to
researchers' intuitions, and formal and informal knowledge. Meanwhile,
algorithms {\em evolve} since those which seem to work well are kept
and varied, and those which do not are thrown away (algorithms not
published or re-implemented, papers not cited, or results not
replicated). Despite this process of specialisation to a problem
subset, and putative better-than-random performance, it is appropriate
to call these algorithms ``generic'' because they are not specialised
to particular application {\em domains} such as engineering design,
VRP, planning, regression, etc., or to individual instances. We are
distinguishing between a problem {\em subset} and a problem {\em
  domain}, as in Section~\ref{sec:undef-terms-probl}.

\citet{yuen2015composing} mention that although ``real-world''
problems do have properties which allow algorithms to out-perform
random search, nevertheless ``the correct lesson to learn from NFL is
that it is {\em unlikely} and probably {\em impossible} to find a
black-box algorithm that would do well on {\em all} real world
problems''. But a theorem that proves a
proposition under certain assumptions does not provide probabilistic
evidence for that proposition if the assumptions are not fulfilled. So
if there is evidence for this position, it is independent of NFL.

This section has explained several specific misunderstandings of NFL,
with speculation as to how they arise. In particular, it has clarified
that generic algorithms are already specialised to a problem subset
and could in principle out-perform random search on all problems of
interest. As argued in Section~\ref{sec:structure}, next, there may be
reason to believe that real-world problems share properties which
would allow a single algorithm to out-perform random search on
all.


\section{Avoiding NFL: Assumptions} \label{sec:structure}

Most problems in the set of ``all possible functions'' on a search
space (as in the original NFL), or in a CUP set (as in SNFL), have
very little structure which can be exploited by search algorithms to
achieve better-than-random performance. But most of the problems we
try to solve using search algorithms do seem to have
structure. \citet{wolpert-macready} write that ``the simple existence
of that structure does not justify choice of a particular algorithm;
that structure must be known and reflected directly in the choice of
algorithm to serve as such a justification.'' In contrast,
\citet{bbcomp} write that ``[i]t is NOT the idea of black box
optimization to solve problems without structure, but rather to
perform well when structure is present but unknown.''
Finally,~\citet{krawiec2009analysis} write of a ``quest for properties
that are common for the real-world problems (or some classes of them)
and that may be exploited in the search process [\ldots]. Examples of
such properties studied in the past include fitness-distance
correlation, unimodality of the fitness landscape, and modularity''.

An algorithm must be specialised to a subset of problems (not a
``problem domain'', as discussed in
Section~\ref{sec:undef-terms-probl}), taking advantage of properties
of that subset, in order to out-perform random search on it. Contrary
to~\citet{wolpert-macready} and \citet{sewell2012forecasting} it is
not necessary to know or state the properties in question: an
algorithm will perform as well as it performs, no matter what the user
knows or states. Many users of generic algorithms which are
specialised to problem subsets achieve good results without being
capable of stating the structural properties of their objective
functions. Indeed many designers of successful generic algorithms have
not stated such properties either.

The types of problems we wish to optimise using black-box methods have
structure which can be formalised in several ways. Well-known
algorithms are already specialised to such structure: thanks to NFL,
we can simply define that an algorithm is specialised to a subset of
problems if it out-performs random search on that subset.
However, identifying the type of structure present in a set of
functions, and just how an algorithm is matched to that set, is the
goal of this section. In each of the following sub-sections, a
different simple property is described which, if it holds across a set
of functions, shows that that set is not CUP, allowing an algorithm to
escape SNFL. Many of the properties which will be identified follow
quickly from the proof by~\citet{igel2001classes} that a non-trivial
neighbourhood on the search space is not invariant under permutation,
hence if we assume that all problems of interest share some property
defined in terms of neighbourhoods then the set of problems of
interest is not CUP. In several cases, it is also identified how
well-known algorithms exploit the structure being discussed.  As
argued in Section~\ref{sec:how-can-performance}, NFL refinements
including FNFL and RMNFL do not add any practical constraint on
algorithm performance to the picture already provided by SNFL, so
escaping SNFL (and thus also the original NFL) is sufficient.  In a
sense, this section responds to~\citet{christensen2001can}, who use
one definition for problem structure, but invite the reader to
operationalise other definitions and use them in the same way.

Problems have their ``natural'' structure only when we use a natural
representation. For example, on the space of bitstrings, the Onemax
problem has strong structure when we use bit-flip as the neighbourhood
operator. We could instead define a different operator, destroying the
structure and making Onemax a difficult problem for typical
algorithms. Similarly, any problem can be converted into something
very similar to Onemax by hand-crafting the neighbourhood operator. In
both cases, such an operator would require a complex description
relative to that of the bit-flip operator. This suggests that we could
formalise the idea of a ``natural'' representation in complexity
terms, and hints at a duality between complexity of objective
functions~\cite{droste1998perhaps, droste2002analysis,
  streeter2003two, whitley2008focused} and complexity of
operators. However, further development of this idea is considered out
of scope. In the following discussion, it will be assumed that
problems are encoded in natural ways.

The properties we will list are alternatives, i.e.~to evade NFL, only
one property needs to hold, not all. The properties are not all
equivalent, in that one may hold, but not another; but more often, a
problem has several. However, observing one of these properties in one
problem instance is clearly not enough to conclude that it is present
in all problems of interest. Instead it is necessary either to show
that the property holds for all considered problems or, as remarked
by~\citet{joyce2018review}, to show it for some and to argue that they
are representative.

\subsection{Locality} \label{sec:self-similarity}
\label{sec:locality}

Strictly speaking, a function is said to have the property of {\em
  locality} if it always maps neighbours to neighbours, for suitable
definitions of neighbourhood in both the domain and range. A looser
idea of locality is common in metaheuristics research: we say that an
objective function has locality if neighbouring points in the search
space have similar objective values. One way to write this is:
$$\Expect_{x\in X} |f(x) - f(N(x))| < \Expect_{x, y \in X} |f(x) - f(y)|$$
\noindent where $N$ is a neighbour function and $X$ is the search
space. That is, the objective value of a pair of neighbours is more
similar than the objective value of a pair of randomly-chosen
points. A correlation or scatterplot between $f(x)$ and $f(N(x))$ will
reveal the presence or absence of such structure in one problem.

If a set of functions have this property, then the set is not CUP, as
demonstrated by~\citet{streeter2003two}. Moreover, this property
directly justifies the use of a neighbourhood (mutation) operator in
an algorithm on this set. Observing this statistical property, and
choosing to use an algorithm with a neighbourhood operator, amounts to
tailoring the algorithm to the problem.


The same idea can be extended to two or more steps of a neighbourhood
operator. A correlation between $f(x)$ and the objective value of the
``neighbour of a neighbour'' $f(N(N(x)))$ also reveals structure on
the search space. An algorithm which uses a neighbourhood operator but
allows for worsening moves, such as simulated annealing, might be said
to exploit such structure.

The analogous property for a crossover operator can be formalised as:
$$\Expect_{x, y\in X}|f(x) - f(C(x, y))| < \Expect_{x, y\in X}|f(x) -
f(y)|$$
\noindent where $C$ is a crossover operator returning one
offspring. Again, \citet{streeter2003two} shows that with this
property, a problem subset is not CUP. This property justifies the use
of an algorithm with a crossover operator: such an algorithm is
tailored to the problem subset. A genetic algorithm is thus already
tailored to the (very large) set of problems with these locality
properties~\citep{serafino2014optimizing}.





\subsection{No maximal steepness} \label{sec:steepness}

An objective function has the property of {\em no maximal steepness}
if the largest function difference between a pair of neighbours is
less than the largest possible function difference. That is:
$$\max_{x\in X, N(x)} |f(x) - f(N(x))| < \max_{x, y\in X} |f(x) -
f(y)|$$
\noindent where $\max_{N(x)}$ is seen as returning the maximum over
all neighbours of $x$.

Given any function, permutation of objective values can give a
function achieving maximal steepness, so this property shows that a
set of functions with this property is not CUP~\citep{igel2001classes,
  igel2004no}. \citet{jiang-chen} name such functions {\em
  discrete-Lipschitz}, by analogy with Lipschitz continuity on
real-valued functions, and use the result to demonstrate cases in
which NFL does not apply. Again, any algorithm that uses a
neighbourhood concept is taking direct advantage of this ``problem
knowledge''.

\subsection{Fitness distance correlation} \label{sec:fitn-dist-corr}

{\em Fitness distance correlation} (FDC)~\citep{jones:forrest}
measures the correlation between $d(x, x^*)$ and $f(x)$, where $x^*$
is the optimum point, $x$ is an arbitrary point, and $d$ is a
distance. In a minimisation problem, large positive FDC values tend to
indicate easier problems and large negative values difficult or
deceptive problems (though counter-examples exist). FDC is thus a
measure of a type of structure in a problem. It can be seen as a
``multi-step'' generalisation of locality.

Given a problem with positive FDC, permutation of objective values can
produce a problem with negative FDC. This is easy to see by picturing
a plot of objective value against distance (a line of best fit through
this data must have a positive slope). By permuting objective values
other than that of the optimum we can achieve a negative slope. Thus,
a set of functions all with positive FDC is not CUP. An algorithm that
tends to visit new points close (in $d$) to good points is tailored to
a subset with positive FDC.


The statistic proposed by~\citet{christensen2001can} has something of
the same meaning as FDC. It is large when for many points, the point's
objective value {\em ranking} is the same as its distance
ranking. They propose to threshold functions according to this
statistic: functions with values above a threshold have structure, and
the set of such functions is not CUP. Moreover, they propose an
algorithm which directly exploits this type of structure.



\subsection{Constraints and penalty functions}

\citet{kimbrough2008feasible} consider NFL in the context of problems
with constraints. One common approach when using metaheuristics is to
add a penalty term for constraint violations to the objective
function. \citeauthor{kimbrough2008feasible} show that if the original
objective $f$ is drawn from a CUP set, but the penalty term $g$ is
fixed and can take on at least two values (e.g.~at least one feasible
and one infeasible point exist), then the composite objective $f -
\lambda g$ is not CUP.




\subsection{Number of local optima} \label{sec:number-local-optima}
\label{sec:unimodality}

{\em Unimodality} is the property that a function has only a single
(global) optimum. This is a strong structure which tends to make
problems easy. A set of unimodal functions is not CUP, and an
algorithm which uses a local neighbourhood operator is specialised to
such a set. \citet{igel2004no} also show that if the maximum possible
number of local optima in the space is not achieved by any of the
functions in a subset, then the subset is not CUP. A generalisation is
possible\footnote{Due to an anonymous reviewer.}: if any fixed number
of local optima (not necessarily the maximum) is not achieved by any
function in the set, then it is not CUP. Even a subset of problems
which includes no unimodal function is not CUP. However, it seems
difficult to describe how an algorithm can be matched to these
properties.

\subsection{Modularity}\label{sec:modularity}

{\em Modularity} is the property that candidate solutions can be
consistently broken down into parts, each of whose contribution to the
objective function is to some extent independent of that of others.
\citet{krawiec2009analysis} propose a measure for modularity which
depends on the degree of monotonicity of a module's contribution to
the objective. Given a problem with some degree of modularity,
permutation of objective values can remove modularity, since it can
remove monotonicity. Therefore, a set of functions each with high
modularity is not CUP. An algorithm which implements a variation
operator by varying just one component of the candidate at a time
(e.g.~a neighbourhood operator which alters just one variable in a
real vector) is aiming to exploit such modularity.

\subsection{Bounded time complexity and bounded description length}
\label{sec:bound-time-compl}
\label{sec:compressibility}

We conclude this section with two types of assumption on problem
structure which have sometimes been proposed as methods of escaping
NFL, but which do not quite work: bounded time complexity and bounded
description length.

In the set of all objective functions, only a very small proportion of
them can run in reasonable time~\citep{droste1998perhaps,
  droste2002analysis}. Objective functions which require unreasonably
long execution time are not realistic candidates for optimisation,
regardless of what performance would in principle be. Therefore, we
may assume that we will attempt to optimise functions which run in
bounded time, and this assumption means our set of functions is much
less than the set of all possible functions. \citet{streeter2003two}
also argues in this direction. However, this argument seems sufficient
only to escape the original NFL, not SNFL or other refinements, since
the counting argument used by~\citep{droste1998perhaps,
  droste2002analysis} considers all possible functions.

In algorithmic information theory, the description length of an object
is the length of the shortest encoding for that object. A function is
{\em compressible} if it has a description length shorter than that of
a lookup table.

\citet{streeter2003two} shows that an NFL result does not hold on a
set of functions if the functions' description length is
``sufficiently bounded''. However, this turns out not to be good
enough for our purposes. \citet{whitley2011no} point out that ``there
is a subset of problems where Best-First local search is likely to be
a useful search method. But there is a corresponding set of functions
where Worst-First local search is equally effective. What do these
functions look like? They probably are not random, but rather
`structured' in some sense''. That is, they have bounded description
length. This is because if they were incompressible, Worst-First could
not do any better than Random Search.

To help us picture such functions, we can use a trap construction:
given a ``nice'' real-world function $f$ on which Best-First search
does well, define $f'$ as $f'(x) = f(x)$, except for a global optimum
$x^+$ and a global ``pessimum'' $x^-$, whose objective values are
swapped: $f'(x^+) = f(x^-)$ and vice versa. On the ``trap'' function
$f'$ a Worst-First searcher can be expected to do well. $f'$ is not
much less compressible than $f$. Importantly, the set $\{f, f'\}$ is
not CUP, but performance of Best-First and Worst-First algorithms will
be equal on it. Thus, observing bounded description length on a set of
problems evades NFL itself, but now the ``almost no free lunch''
(ANFL) theorem constrains
performance. ANFL~\cite{droste2002optimization} shows that assuming
low complexity in the problems of interest is not sufficient: for
every problem $f_1$ where our algorithm $A$ out-performs RS, there is
another $f_2$ of similar complexity where RS out-performs
$A$~\cite{joyce2018review}
.

\citet{droste2002analysis} give a more rigorous treatment on
constructing these difficult, but not unstructured problems. Although
bounded time complexity or description length do not escape NFL
results, the nature of their construction, and of the inversion
example above, may give us comfort that we will not encounter such
functions as real problems very often.

This section has examined {\em abstract} structure which may be
present in problems. Next, several {\em concrete} NFL counter-examples
are demonstrated.


\section{Avoiding NFL: Examples} \label{sec:examples}

In this section mechanisms and strategies for avoiding NFL are
demonstrated by example. Several are well-known as corollaries to
previous work: our goal is to walk through the reasoning.

\subsection{MAX-2-SAT is not CUP} \label{sec:max-2-sat}

\citet{whitley2008focused} remark that previous work
by~\citet{igel2001classes} and by~\citet{streeter2003two} has shown
that MAX-SAT is not CUP; however the result is not stated
explicitly. We will present a MAX-2-SAT problem whose objective
value-permutation is not a MAX-2-SAT problem, showing that MAX-2-SAT
is not CUP.  Let us consider the search space of $n=3$ variables and
the instance defined by the formula $\phi = (x_0 \vee x_1)$. We will
order points in the search space in the natural way, and for each
calculate its objective value, giving an objective table as shown in
Table~\ref{tab:max-2-sat-phi}.



\begin{table}
  \centering
  \caption{Objective-value table for MAX-2-SAT with $n=3$ and the
    formula $\phi = (x_0 \vee x_1)$. $f(x)$ is the number of clauses
    of $\phi$ satisfied by $x$. \label{tab:max-2-sat-phi}}
\begin{tabular}{ccc|c}
  $x_0$ & $x_1$ & $x_2$ & $f(x)$ \\
  \hline
  0 & 0 & 0 & 0 \\
  0 & 0 & 1 & 0 \\
  0 & 1 & 0 & 1 \\
  0 & 1 & 1 & 1 \\
  1 & 0 & 0 & 1 \\
  1 & 0 & 1 & 1 \\
  1 & 1 & 0 & 1 \\
  1 & 1 & 1 & 1 \\
\end{tabular}
\end{table}


We now permute the objective values to (0, 1, 1, 0, 1, 1, 1, 1). We
will see that the latter cannot arise as the MAX-2-SAT objective-value
table of any formula $\phi'$ on the same number of variables (not
necessarily the same number of clauses). The first entry ($0, 0, 0
\rightarrow 0$) implies that $\phi'$ contains no clauses featuring
$\neg x_0$, $\neg x_1$, or $\neg x_2$ whatsoever; the last entry ($1,
1, 1 \rightarrow 1$) implies that $\phi'$ contains no more than one
clause featuring any of $x_0$, $x_1$, $x_2$. Together, these imply
that the formula must consist of a single clause, composed of $x_i$
variables only (no negations). Therefore there are only three
possibilities: $\phi' = x_0 \vee x_1$, $\phi' = x_0 \vee x_2$, or
$\phi' = x_1 \vee x_2$. None of these match the rest of the
table. Therefore, no such $\phi'$ exists, and so the set of MAX-2-SAT
problems on 3 variables is not CUP.

It is interesting to see that MAX-2-SAT achieves maximal steepness
(see Section~\ref{sec:structure}). For example, in the 4-variable
problem defined by the target formula $(x_0\vee x_1) \wedge (x_0\vee
x_2) \wedge(x_0\vee x_3)$, the minimum objective value is 0 and the
maximum is 3, and the two neighbours (0, 0, 0, 0) and (1, 0, 0, 0)
achieve these values. As discussed in Section~\ref{sec:structure}, if
a problem does not achieve maximal steepness, then it must not be
CUP. MAX-2-SAT does achieve maximal steepness, but as presented here,
a different argument shows that it is still not CUP. Thus, there may
be multiple routes to showing that SNFL does not apply.

\subsection{TSP is not CUP} \label{sec:tsp-not-cup}

\citet{koehler2007conditions} prove, using the idea of {\em circulant
  matrices}, that symmetric TSP is not CUP. \citet{jiang-chen} show
that TSP instances are discrete-Lipschitz, and thus not CUP (see
Section~\ref{sec:steepness}). We wish to go further by constructing a
concrete counter-example, that is a TSP which, when its objective
values are permuted, is not a TSP -- showing that SNFL does not apply
to TSP. Note that although the following example takes advantage of
the fact that the {\em problem} is not black-box, in order to
construct a counter-example to NFL, it is not avoiding NFL by using a
non-black box {\em algorithm}.

\begin{figure}
  \centering
  \subfigure{
    \includegraphics[width=0.425\linewidth]{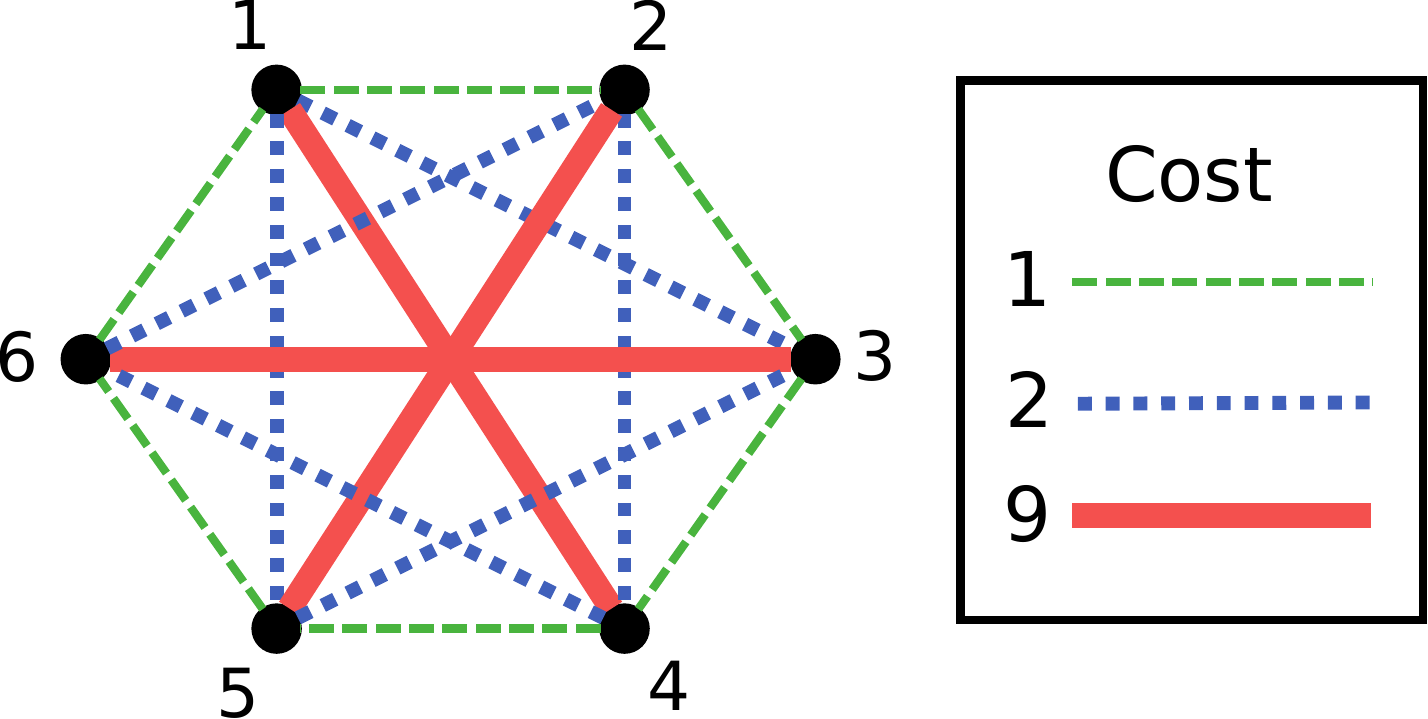}
  }
  ~~~~~~~~~~
  \subfigure{
    \raisebox{0.5\totalheight}{
      $\begin{bmatrix}
        0 & 1 & 2 & 9 & 2 & 1 \\
        1 & 0 & 1 & 2 & 9 & 2 \\
        2 & 1 & 0 & 1 & 2 & 9 \\
        9 & 2 & 1 & 0 & 1 & 2 \\
        2 & 9 & 2 & 1 & 0 & 1 \\
        1 & 2 & 9 & 2 & 1 & 0
      \end{bmatrix}$
    }
  }  \hfill
  \caption{A non-geometric TSP instance shown graphically and as a
    cost matrix.\label{fig:tsp}}
\end{figure}

Figure~\ref{fig:tsp} shows a TSP problem on 6 cities. The best tour
$x=(123456)$ has $f(x) = 6$; the worst $x'=(142536)$ has $f(x') =
32$. To construct a counter-example we {\em permute the objective
  value of the best and worst solutions}, that is we let $f(x) = 32$
and $f(x') = 6$, and show that the resulting $f$ is not a TSP. The
idea we are trying to exploit is that in many problems, the neighbours
of the optimum are likely quite good, and similarly the neighbours of
the worst individual in the space are likely to be bad. This is an
approximate paraphrase of the main idea of {\em maximal steepness}. We
observe that in the new problem, the worst individual $x=(123456)$ has
objective value $32$ and has 6 neighbours under 2-opt mutation, all
with objective value $8$, e.g.~$f(123465) = 8$. Is there any cost
matrix that could give a problem with these objective values? We
designate the new, unknown cost matrix as $C$. These properties give
us 7 simultaneous equations in the coefficients $C_{ij}$:

\begin{align*}
  C_{12} + C_{23} + C_{34} + C_{45} + C_{56} + C_{61} &= 32\\
  C_{12} + C_{23} + C_{34} + C_{46} + C_{65} + C_{51} &= 8\\
  C_{12} + C_{23} + C_{35} + C_{54} + C_{46} + C_{61} &= 8\\
  C_{12} + C_{24} + C_{43} + C_{35} + C_{56} + C_{61} &= 8\\
  C_{13} + C_{32} + C_{24} + C_{45} + C_{56} + C_{61} &= 8\\
  C_{13} + C_{34} + C_{45} + C_{56} + C_{62} + C_{21} &= 8\\
  C_{16} + C_{62} + C_{23} + C_{34} + C_{45} + C_{51} &= 8
\end{align*}

\noindent Using manual methods or a computer algebra system (a link to
code is given later) we will see that these equations have no
solution. Thus, TSP on 7 cities is not CUP. It is the physical
structure of the problem -- the objective is the sum of inter-city
distances -- that escapes NFL.

\subsection{Symbolic Regression is not CUP} \label{sec:symb-regr-not}

\citet{poli2009free} use a nice geometric argument to show that
genetic programming symbolic regression (GPSR) is not CUP, that is
that permuting fitness values of a GPSR problem can give a new problem
which is not an instance of GPSR. Although GPSR is a supervised
machine learning method, here we are discussing NFL for search and
optimisation, which is about search (training) performance only,
rather than NFL for supervised machine learning, which is about
performance on unseen data only.

\begin{figure}
    \centering
    \includegraphics[width=0.3\linewidth]{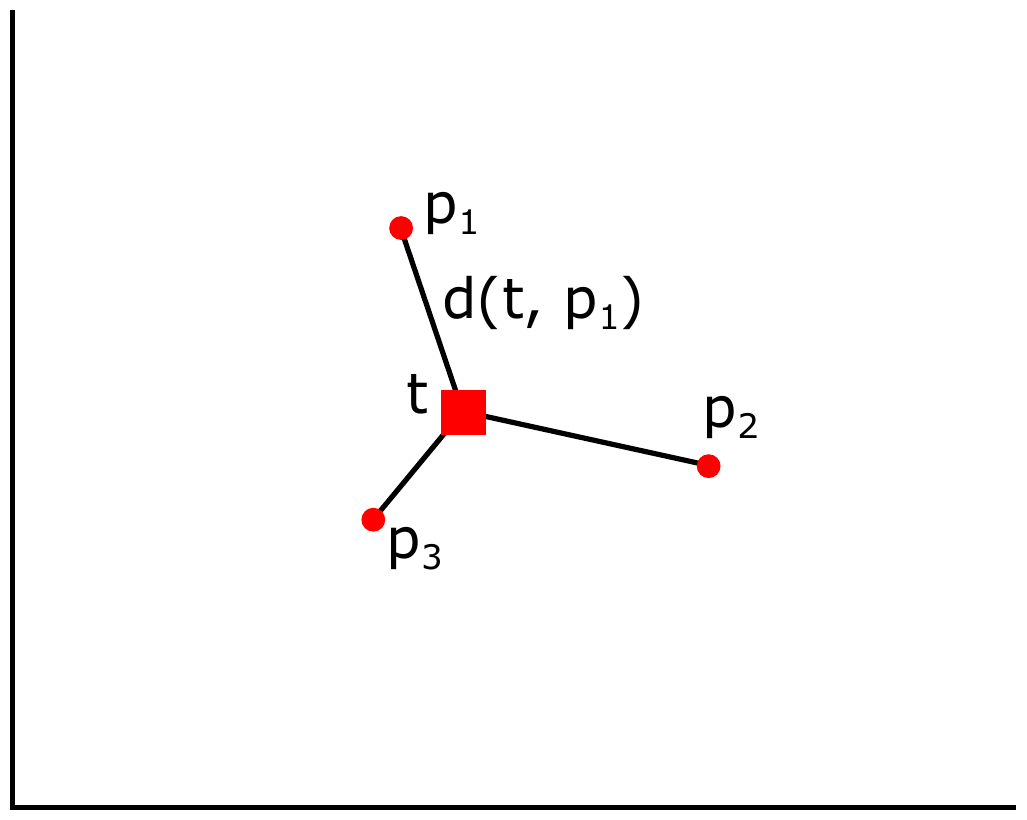}
    \hfill
    \includegraphics[width=0.3\linewidth]{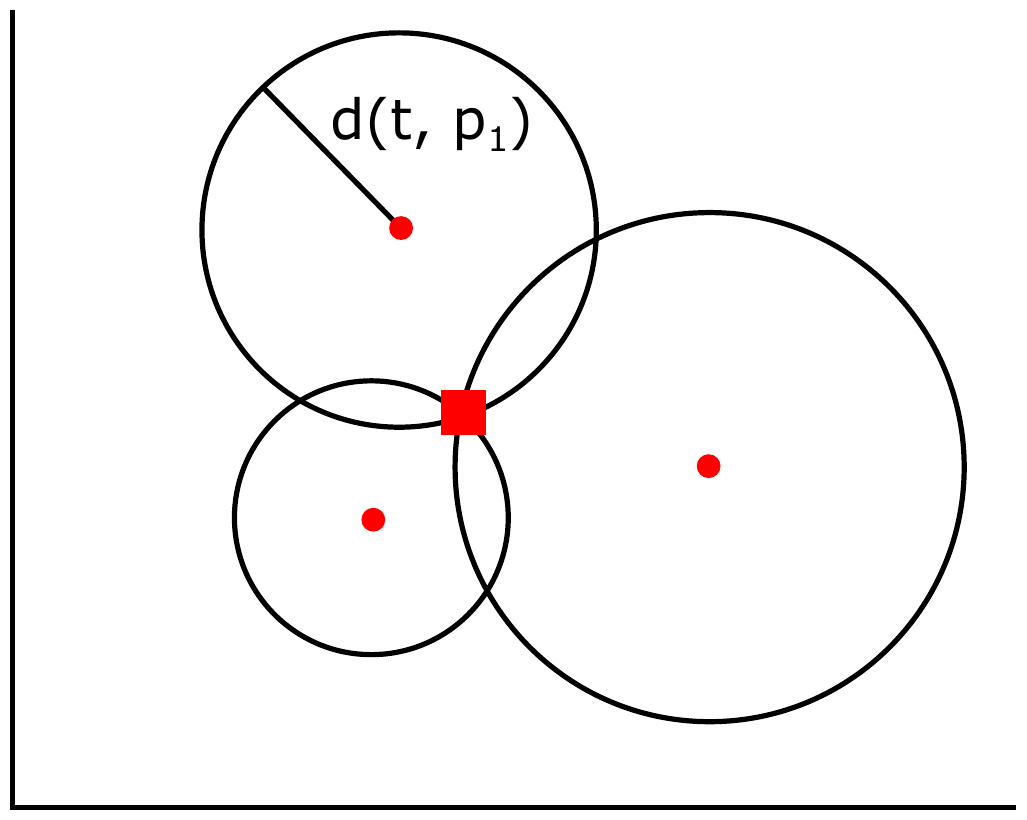}
    \hfill
    \includegraphics[width=0.3\linewidth]{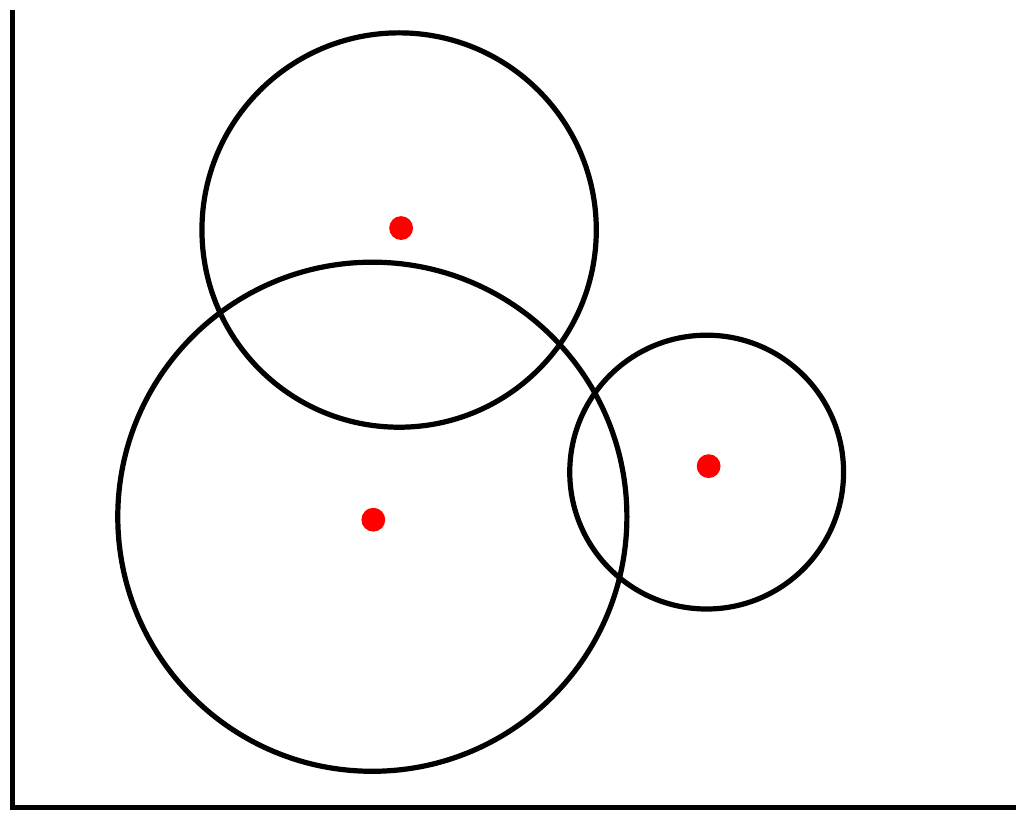}
 \caption{Symbolic regression is not CUP, as shown
   by~\citet{poli2009free}. Left: a GPSR problem in semantic
   space. Each axis indicates program output on one objective function
   case. Points $p_i$ indicate semantics of various programs. The
   objective function is RMSE, i.e.~Euclidean distance in semantic
   space from the target semantics $t$. Centre: $t$ is at the
   intersection of hyperspheres with centres $p_i$ and radii $f(p_i) =
   d(t, p_i)$. Right: after permutation of objective values among
   programs, there is no such intersection, so the problem is not a
   symbolic regression problem for any target semantics
   $t$.\label{fig:tsp-poli}}
  \end{figure}

The argument is briefly summarised in Fig.~\ref{fig:tsp-poli}. It
takes advantage of the fact that GPSR is not really a black-box
problem: instead, the objective function is a function of a sum over
partial objectives, one per item in the training data.

Another issue relevant to NFL and GPSR is duplicated
semantics. Consider any space which includes the two functions {\tt (*
  x 2)} and {\tt (+ x x)}, where {\tt x} is a variable, and includes
other functions also. These two functions are distinct items in the
search space, so under permutation they can get distinct objective
values. But if these two trees, which are semantically identical, get
distinct objective values, then the objective function is not symbolic
regression. Therefore the set of SR problems on a fixed search space
is not CUP.
It is not that over-representation (multiple programs with the same
semantics) is a technique that helps GP to perform well: rather, it
just means that SNFL does not apply.

\subsection{Boolean genetic programming is not CUP}

A similar argument can be made for Boolean genetic programming, for
example, where now target semantics is a binary vector, and $d$ is
Hamming distance. Although the argument is similar, distance on the
page does not give a reliable intuition for $d$, so a concrete example
is worthwhile. With two variables $x_0$ and $x_1$, we can define
example programs such as $p_0(x) = x_0 \wedge x_1$, $p_1(x) = x_0 \vee
x_1$, and $p_2(x) = \neg x_0$. Taking a target semantics $t = (0, 0,
0, 1)$, our programs receive objective values $f(p_0) = 4$ (because
$p_0 = x_0 \wedge x_1$ has exactly the semantics $(0, 0, 0, 1)$),
$f(p_1) = 2$, $f(p_2) = 1$, which we write as just $(4, 2,
1)$. Permuting these objective values to e.g.~$(4, 1, 2)$, we find
there is no target semantics $t$ which would give these three programs
these three objective values.

Code for exploring these examples (TSP, MAXSAT, and Boolean GP)
is available from \url{https://github.com/jmmcd/NFL}.






\section{The Anthropic Principle and NFL} \label{sec:discussion}
\label{sec:other-nfl}
\label{sec:real-meaning-nfl}

In the previous two sections we have given several specific mechanisms
by which researchers can ``escape'' NFL results. However, this paper
is arguing for a stronger position: researchers can in many common
situations ignore NFL without specific, per-paper justification. In
this section, we broaden the discussion to include ``the other NFL'',
NFL for machine learning (NFLML). We then consider both together and
propose an {\em a priori} assumption which justifies ignoring NFL in
many common situations.

NFLML~\citep{wolpert1996lack} states that any two supervised machine
learning algorithms achieve the same performance on unseen data,
averaged over all possible problems. It has a very similar flavour to
NFL for search and optimisation, but was less controversial in its
field, perhaps partly because the ideas had been more
anticipated. \citet{schaffer1994conservation} proved the central
result, but did not use the NFL nickname. Both NFL for search and
optimisation and NFLML may be said to have roots in the ``algorithm
selection problem'' posed by~\citet{rice1976algorithm}, on which
research continues, e.g.~considering a meta-learning
approach~\citep{smith2009cross}.


\citet{schaffer1994conservation} comments on a common attitude
concerning NFLML, that it is ``theoretically sound, but practically
irrelevant''. The position is summed up by~\citet{domingos2012few} in
comments which echo those often made concerning NFL for optimisation:
``very general assumptions -- like smoothness, similar examples having
similar classes, limited dependences, or limited complexity -- are
often enough to do very well''. However,
\citet{schaffer1994conservation} also cites real-world examples in
which ML algorithms have indeed performed worse than random on unseen
data. Such cases have been observed also in NFL for search and
optimisation. However, in both cases the same evidence -- worse than
random performance -- can arise even in cases where NFL does not
strictly apply.

Researchers sometimes misunderstand NFLML in a similar way to NFL, for
example~\citet{smith2009cross} states that the StatLog project
``confirmed that no single algorithm performed best for all problems,
as supported by the [NFLML] theorem'', even though no NFLML result
can apply to the set of problems mentioned. The distinction between
problem subsets and application domains is often blurred in the
context of NFLML, just as in NFL for search and optimisation (see
Section~\ref{sec:undef-terms-probl}), for example in the NFL
discussion in~\citet{murphy2012machine}, p.~24.

Schaffer also pointed out that NFLML is a formalisation of the basic
problem of induction -- drawing any generalisation from observed data
is impossible without an additional assumption or bias -- and dates
this to~\citet{hume-treatise}. In fact, according
to~\citet{sewell2012forecasting}, NFLML ``formalizes Hume, extends him
and calls all of science into question''.  We cannot learn (in the
machine learning sense, or in the scientific sense of proceeding from
observations to principles and predictions, or even in the everyday
sense) without a suitable bias, and it is claimed that there is no
{\em a priori} justification for choosing any
bias. \citet{serafino2013no} also links NFL for search and
optimisation to the problem of induction.


And yet, our learning algorithms do learn. The reasons that these
types of learning work (despite Hume and NFLML) and metaheuristic
search works (despite NFL for search and optimisation) are really the
same reason, and we now suggest that the real value of NFL is that it
forces us to identify that reason. More specifically, why do the
problems which seem to commonly arise have the property that simple,
generic algorithms can out-perform random search, even though such
problems are a tiny fraction of the {\em possible} problems?

Our universe runs on fairly simple rules. ``[T]he class of functions
of practical interest can be approximated through `cheap learning'
with exponentially fewer parameters than generic ones, because they
have simplifying properties tracing back to the laws of physics. The
exceptional simplicity of physics-based functions hinges on properties
such as symmetry, locality, compositionality and polynomial
log-probability''~\citep{lin2017does}. This comment was made in the
context of supervised learning-type functions, but the reasoning holds
for optimisation objective functions also. If the universe is simple
and rule-bound, so that unseen data is in some way similar to training
data, then the processes that generate fitness landscapes on real
problems will usually be simple and rule-bound too.

Even if we do not know physical laws precisely, we know them
approximately, and this is sufficient to know that our universe has
exploitable structure. As stated by~\citet{hutter2010complete}, ``The
assumption that the world has some structure is as safe as (or I think
even weaker than) the assumption that e.g.~classical logic is good for
reasoning about the world''. So we may ask: what would a universe
without exploitable structure look like? \citet{wolpert2012no}
describes a scenario (two professors competing to produce good ML
algorithms) which illustrates it. In a universe without structure, no
amount of evidence in favour of one professor's algorithms could
justify betting that that professor would continue to be the
best. \citet{wolpert2012no} argues that to resolve this, and generally
to justify proceeding from evidence to predictions, it is required to
make an assumption about the ``probability distribution over
universes'', which cannot be justified {\em a priori}.

However, there is a well-known position, not previously considered in
this context, which justifies exactly such an assumption: the {\em
  anthropic principle}. It states that ``the Universe (and hence the
fundamental parameters on which it depends) must be such as to admit
within it the creation of observers within
it.''\citep[][p.~294]{Carter1974}. If it were not, observers would not
be here to observe otherwise.


Any intelligent organism makes predictions about the future and seeks
to act on them; any behavioural organism acts on the basis of implicit
predictions about the future; and any organism at all embodies an
implicit prediction about the environment it will find itself in. If
these predictions are systematically wrong, these organisms are less
likely to survive, propagate, and evolve. Any organism requires on
some robustness in these things, since there will always be noise in
the genetic copying procedure and in the environment sufficient to
make the outcome differ from the ideal. We are assuming here that
organisms arise through evolutionary processes involving selection and
a copying procedure which leads to variation.
\citet{haggstrom2007uniform} remarks that the type of search landscape
we observe in biology is highly ``clustered'' or auto-correlated -- a
single mutation in the DNA of a surviving, reproducing organism does
not (in expectation) result in an outcome as bad as generating DNA
uniformly from scratch. One aspect of the explanation for this is that
gene interactions are not so strong as to overwhelm an overall
additive behaviour of fitness in response to
genes~\cite{obolski2017key}. If biological search landscapes were not
structured, then evolution would not work at all: intelligent life
could not arise. The professors of Wolpert's scenario cannot evolve to
exist in the universe he describes.

\citet{mendes2004fully} wonders whether structured problems turn out
to be common just because the universe is rule-bound, or because they
are salient (i.e., of interest) to observers. In fact these
possibilities are really the same possibility, since as argued
observers evolve to take advantage of the rules of the universe.

Solomonoff induction~\citep{solomonoff1964formal}
similarly allows for a principled approach to induction (justifying
scientific enquiry and machine learning, {\em contra} Hume and NFLML)
by assuming that processes governed by short Turing machines are more
likely to occur (e.g.~as ML problems) than ones governed by long
ones. This assumption is justified {\em a priori} by the anthropic
principle.


Thus, the anthropic principle gives us an {\em a priori} assumption
about the distribution of universes we could find ourselves in, which
allows us to escape NFL: we can assume that our problems of interest
are ones which arise and are salient to observers in a rule-bound
universe. Everyday learning works, science works, supervised learning
works, and metaheuristic search works, {\em because we are here}.




\section{Conclusions} \label{sec:conclusions}

We have observed a collection of evidence that NFL is often
misunderstood in the literature. In response we have stated several
sets of accessible arguments, both new and old. We have argued against
one common position on NFL -- that in order to out-perform random
search, algorithms need to be intentionally tailored to specific
problems -- and for the position that the anthropic principle
justifies {\em a priori} ignoring NFL in many common situations.

\label{sec:things-that-remain}

Many of the practical lessons sometimes stated to follow from NFL are
in fact independent of it, but may still represent excellent advice:

\begin{itemize}
\item There probably isn't one algorithm that wins on all real-world
  problems. As argued in Section~\ref{sec:algorithm-must-be}, NFL
allows the scenario that one algorithm (even one already in existence)
is better than random search on {\em all} real-world problems. But
empirical evidence is obviously against it. It seems likely that
researchers will remain in ``full employment''~\citep{neri2012primer}
for now.



\item Specialising an algorithm with domain-specific or problem subset
  knowledge often helps. ``[A]pplying a general purpose `black box'
  search algorithm is wasteful''~\citep{bbcomp}.  Among many others,
  \citet{bonissone2006evolutionary} argue for this position using both
  NFL reasons and NFL-independent examples.

\item Random search is often a worthwhile baseline, not least because
  it is simple to implement. Out-performing it seems essential for a
  publishable result. This remains true even where NFL is known not to
  apply.


\end{itemize}



We have not yet seen any example of a set of objective functions which
arise as real-world problems and which are either ``all possible
objective functions'' for a given search space, or are closed under
permutation. Neither have we seen a similar example of a focus set
(FNFL) or restricted set (RMNFL). Therefore, the evidence of decades
of research suggests that the burden of proof is on those who claim
that NFL has practical relevance. To be specific: when and why can
researchers ignore NFL?

A researcher may observe better than random performance of an
algorithm on a set of test problems in a real-world problem domain,
and wish to draw a conclusion about performance on new problems of the
same or different sizes, or drawn from different problem domains. As
discussed in Section~\ref{sec:misund-theor}, NFL can be ignored, but a
claim that performance will generalise still requires support.

It is unlikely that a researcher will specifically wish to make a
claim about algorithm performance averaged over the set of all
possible problems on a space (original NFL), a CUP set (SNFL), focus
set (FNFL) or restricted set (RMNFL), but clearly in such a case NFL
cannot be ignored and no improvement over random search is possible.

When a researcher can state an assumption on the structure of their
problem of the type discussed in Section~\ref{sec:structure}, or
follow a template like those in Section~\ref{sec:examples}, then NFL
will not constrain algorithm performance, but of course in doing so
the researcher is taking account of NFL, not ignoring it.

A researcher may prefer not to deal with such assumptions
per-problem. The anthropic principle (as discussed in
Section~\ref{sec:discussion}) justifies an assumption that structure
(often of the types identified in Section~\ref{sec:structure}) will be
present across sufficiently many of the problems of interest for
generic, well-known algorithms to be sufficiently specialised (as
discussed in Section~\ref{sec:misund-theor}) to out-perform random
search on average. As long as a researcher restricts attention to
problems of interest they can use the anthropic principle and ignore
NFL.

A researcher may wish to aim for a ``super algorithm'', one that is
better than random {\em on all real-world problems}. A researcher may
even hope that an existing algorithm is such a super
algorithm. Although it seems unlikely, no NFL result prevents this and
such a researcher can ignore NFL. A researcher who wishes for a
``super algorithm'' better than random {\em on all problems} is
thwarted by NFL.

Of course, nothing in this paper is intended to suggest that
researchers can simply {\em assume} good performance, or good
generalisation. Neither should researchers take advantage of any NFL
discussion to reinvent old algorithms disguised by novel metaphors and
supported by dubious experimental evidence -- the type of behaviour
identified by~\citet{sorensen2015metaheuristics} and
\citet{weyland2015critical}.

For future work, probably the biggest genuine research gaps for those
focussing on NFL itself are (1) characterisation of the focus sets and
restricted sets of FNFL and RMNFL and variants, (2) further study of
how exactly assumptions of structure present in problems are embodied
in new and existing algorithms, and (3) further characterisation of
the grey area beyond what are currently known to be problems of
interest.


\section*{Acknowledgements}
Thanks to the reviewers for suggesting some
significant improvements. This work was carried out while the author
was at University College Dublin.

\bibliographystyle{unsrtnat}
{\small
\bibliography{bibliography}
}

\end{document}